\documentclass[letterpaper]{article} 
\usepackage[]{aaai23}  
\usepackage{times}  
\usepackage{helvet}  
\usepackage{courier}  
\usepackage[hyphens]{url}  
\usepackage{graphicx} 
\usepackage{natbib}  
\usepackage{caption} 
\usepackage{algorithm}
\usepackage{algorithmic}

\usepackage{newfloat}
\usepackage{listings}
\DeclareCaptionStyle{ruled}{labelfont=normalfont,labelsep=colon,strut=off} 
\floatstyle{ruled}
\newfloat{listing}{tb}{lst}{}
\floatname{listing}{Listing}
\usepackage{rotating}
\usepackage{subcaption}
\usepackage{xcolor}
\usepackage{amsmath}
\usepackage{amsfonts}

\definecolor{darkgreen}{cmyk}{1., 0., 1., 0.5}

\title{Robustness in Fatigue Strength Estimation}
\author{
    Dorina Weichert,\textsuperscript{\rm 1, \rm 4}
    Alexander Kister, \textsuperscript{\rm 2}
    Sebastian Houben, \textsuperscript{\rm 3}
    Gunar Ernis, \textsuperscript{\rm 1}
    Stefan Wrobel \textsuperscript{\rm 1, \rm 4}
}
\affiliations {
    \textsuperscript{\rm 1} Fraunhofer Institute for Intelligent Analysis and Information Systems IAIS\\
    \textsuperscript{\rm 2} German Federal Institute for Materials Research and Testing, Section S.3 eScience\\
    \textsuperscript{\rm 3} University of Applied Sciences Bonn-Rhein-Sieg\\
    \textsuperscript{\rm 4} Fraunhofer Center for Machine Learning\\
    {\{}dorina.weichert, gunar.ernis, stefan.wrobel{\}}@iais.fraunhofer.de, alexander.kister@bam.de, s.houben@h-brs.de
}

\begin{document}

\maketitle

\begin{abstract}
Fatigue strength estimation is a costly manual material characterization process in which state-of-the-art approaches follow a standardized experiment and analysis procedure. In this paper, we examine a modular, Machine Learning-based approach for fatigue strength estimation that is likely to reduce the number of experiments and, thus, the overall experimental costs. Despite its high potential, deployment of a new approach in a real-life lab requires more than the theoretical definition and simulation. Therefore, we study the robustness of the approach against misspecification of the prior and discretization of the specified loads. We identify its applicability and its advantageous behavior over the state-of-the-art methods, potentially reducing the number of costly experiments.
\end{abstract}

\section{Fatigue Strength Estimation: Setting}
Fatigue strength is a material property that describes the maximum load that can be applied to a defined specimen for a number of cycles that is thought of as an infinite lifetime. 

More precisely, we distinguish between the fatigue strength of the material and the fatigue strength of specimens from the material: the fatigue strength of the material is the distribution of the fatigue strengths of the specimens from this material. Following \citet{DIN}, we assume that the fatigue strength of the material is log-normally distributed. We denote its mean by \(\mu_L\) and its standard deviation by \(\sigma_L\).
The specimen’s fatigue strength is not a directly observable variable. If a load is applied, we can only measure if the specimen breaks as its fatigue strength is smaller than the load (a so-called failure); or if it survives as its fatigue strength is larger than the load (a so-called runout).

To determine the parameters of the material’s fatigue strength distribution, fatigue testing aims to generate a valid statistic of runouts and failures.
As each experiment is costly (approximately 10,000~\$ and two months), it is imperative to have a very efficient testing procedure that does not generate experiments that cannot be used at analysis time.

\begin{figure*}
    \centering
    \includegraphics{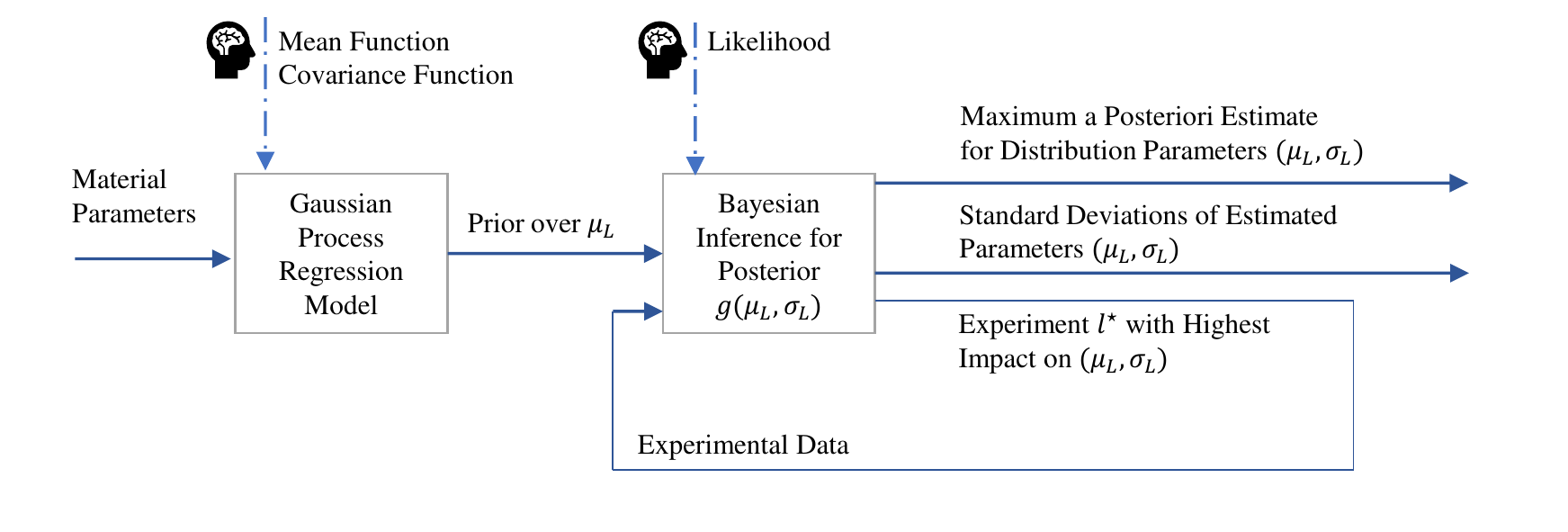}
    \caption{Modular fatigue strength estimation approach.}
    \label{fig:experimental_procedure}
\end{figure*}
State-of-the-art approaches for fatigue testing, such as the current standard \cite{DIN}, are restricting test protocols with difficult to define hyperparameters. These hyperparameters have a massive impact on the experimental efficiency: their misspecification results either in experiments that cannot be used for the analysis or in a large number of experiments required to estimate the fatigue strength parameters. In practice, these hyperparameters are defined by the process engineers using their expert knowledge, thus strongly depending on their individual experience.

This paper describes an alternative approach that combines historical experimental data and the process engineers' expert knowledge. This approach is two-stage (see figure \ref{fig:experimental_procedure}): in the first module, a Gaussian Process Regression Model (GP) merges historical data with expert knowledge to create a prior distribution over the mean fatigue strength \(\mu_L\) of a new material. In the second module, traditional Bayesian Inference generates a posterior estimate of the mean fatigue strength given an actual experimental series.

The most advantageous characteristic of this modular approach is that it is hyperparameter-free for the operating process engineer, thus avoiding potentially costly misspecifications. Additionally, the Bayesian Inference module provides three opportunities:
\begin{itemize}
    \item to generate a maximum a posteriori (MAP) estimate of the materials' fatigue strength parameters (\(\mu_L, \sigma_L\)) after each experiment,
    \item to generate an estimate of the uncertainty of the MAP estimate by calculation of the posterior variance, and
    \item to derive acquisition functions for generating new data, thus defining a testing protocol.
\end{itemize}

To be useful in practice, the testing protocol has to fulfill additional robustness requirements apart from an advantageous convergence behavior. In this paper, we study two of them: Firstly, the Bayesian Inference module's robustness against its prior misspecification. This case can especially occur in the out-of-sample use of the GP, e.g., creating predictions for an aluminum alloy or a cast iron instead of stainless steel, which was used as training data. Secondly, we study the robustness of the approach against rounding of the loads recommended by the acquisition function. In practical use in the lab, the experimental setup is limited to loads from the discrete space - therefore, we experiment with loads rounded to powers of ten.

In a nutshell, our contributions are as follows: we define a new mathematically sound testing and analysis protocol for fatigue strength that incorporates historical data and expert knowledge. Then, we show its practical use by examining its robustness against prior misspecification or discretization issues in a convergence study, benchmarking against a state-of-the-art testing protocol.

\subsection{The Staircase Method for Fatigue Strength Estimation}
The current standard \cite{DIN} describes multiple data analysis methods but only one test protocol for fatigue strength estimation. This protocol is an iterative experimental procedure, called the staircase method. This method assumes the independence of the fatigue strength of different materials, e.g., expecting no correlation of the fatigue strength for steels with similar characteristics.

As we use the staircase method for benchmarking our approach, we quickly revisit its working principle: 
Before applying the staircase method, the process engineer defines load levels for experimentation. 
These load levels are defined by \(L_i = L_{ini} \cdot d^i\), 
where \(L_{ini}\) is a user-defined initial load, \(d\) is the user-defined step size, and \(i \in \mathbb{Z}\). 

Given the load levels, experimentation follows a strict protocol: if a specimen at a specific load is a failure, the load level for the next experiment is reduced by one step; if it is a runout, the load level is raised by one step. 

For later analysis, the generated experimental series has to fulfill multiple requirements: in a valid series, the initial load level is reached at least once again during experimentation. Additionally, the series must contain at least three load levels and at least two turning points where a runout is followed by a failure or the other way around. 

To match these conditions, the parameters \(L_{ini}\) and \(d\) have to be chosen carefully: if \(L_{ini}\) is far away from the ground truth value, a large set of experiments is wasted reaching loads of interest; if \(d\) is too large, it is possible to have only two load levels, not three; if \(d\) is too small, many experiments are necessary to gain two turning points.

Given a full series of multiple experiments, the mean fatigue strength is estimated using two parameters: the lowest valid load level \(L_0\) and the number \(l_k\) each load level was reached, where \(k \in \mathbb{N}_0\) and \(k = 0\) refers to the lowest valid load level. Please note that only in seldom cases \(L_0 = L_{ini}\): the initial load can be higher than \(L_0\), or so small that it was cut from the experimental series to match the analysis requirements.
We then find:
\begin{equation}
    \mu_L = L_0 \cdot \frac{\sum_k k \cdot l_k}{\sum_k l_k}~.
\end{equation}
The standard deviation \(\sigma_L\) is calculated using a more complex heuristic \cite{DIN}. In practice, the reliable estimation of this value requires many experiments and is rarely performed. Even though our testing protocol can estimate it, we will not study its convergence behavior as it is of limited practical use.

\subsection{Related Work}
Apart from the standardized approaches for fatigue testing, several authors have put substantial effort into generating machine learning models predicting fatigue strength from material parameters \cite{agrawal2014,chen2022fatigue,He2021,HE20212,Schneller2022,Shiraiwa2017,Wei2022,xiong2020machine}. Our approach differs from those as we make use of a machine learning model inside of a new experimental protocol, instead of solely making a prediction. Apart from that, our model is of Bayesian type, taking into account available expert knowledge of the process engineers when defining the prior.

\citet{Ling2017HighDimensionalMA} define a framework named FUELS for material discovery that builds on Bayesian Optimization and demonstrates its use for finding new materials with defined fatigue strength. In contrast, our work concentrates on characterizing, not designing, a new material and support the process engineers in carrying out their experiments more efficiently.

Similar to our approach is the recent work of \citet{BO_fatigue} published as a preprint. Here, the authors define a similar posterior over the fatigue strength parameters (\(\mu_L, \sigma_L\)). Our work differs in two main aspects: firstly, their prior is built on simulations of perturbing a well-known heuristic, thus not considering the correlation of fatigue strength of different materials. Secondly, their acquisition function is limited in the staircase manner to predefined load levels, thus reproducing the problems of the staircase approach: it potentially generates superfluous experiments in case of misspecified hyperparameters.

\section{Bayesian Inference for Fatigue Strength Estimation}
Arising from practical considerations, a new fatigue strength testing, and analysis protocol has to fulfill several requirements. First, it should be free from hyperparameters that potentially degrade the test efficiency in the case of misspecification. 
Second, the approach should consider the correlation of different materials' fatigue strength based on historical data and expert knowledge. The very conservative independence assumption of the staircase method is justified for entirely new and unknown materials. However, in the application area of stainless steels, the correlation has already been exploited in machine learning models \cite{agrawal2014,chen2022fatigue,He2021,HE20212,Schneller2022,Shiraiwa2017,Wei2022,xiong2020machine}. Third, all available expert knowledge, i.e., about the experimental setup and behavior, should be taken into account when defining the experimental procedure.

We formulate a modular approach for efficient fatigue strength estimation to meet all these requirements; see figure~\ref{fig:experimental_procedure}. 

\subsection{Gaussian Process Regression Model}
The first module consists of a GP trained on historical data that expresses the correlation of the mean fatigue strength \(\mu_L\) for different materials. This predictive model differs from recent machine learning approaches as we use a Bayesian, not a heuristic model, so the prediction is a normal distribution over \(\mu_L\). 
The used training data was offered from a partner company and is based on the fatigue data from 114 stainless steels. Unfortunately, we are not allowed to publish the data and thus the final trained model, as the GP offers direct access to the train data.
Our model uses four relevant dimensions of an experiment as input: the loaded volume V90 of the specimen, the specimen's edge hardness, the load type (e.g., bending, stress and strain), and the load ratio R, which describes the ratio between the maximum and the minimum load amplitude. The output for training was the related mean fatigue strength \(\mu_L\) estimated via historical experiments.   

Before training the model, we make a train-test split of size 80/20. 
To cope with the data's log-normality, we logarithmize the mean fatigue strength data before applying standardization. 

Afterward, we use a constant zero mean function and a tailored covariance function \(k_{L}\) that was defined with the help of process engineers. This covariance function merges the assumption of a linear trend with an expected very smooth local behavior by summing a linear covariance function \(k_{lin}\) with automatic relevance determination with a rational quadratic covariance function \(k_{RQ}\). 
Thus, our covariance function is defined as follows:
\begin{equation*}
    \begin{split}
        k_{L}(x, x') &= k_{lin} + k_{RQ} \\ 
        &= \sum_{d=1}^D \sigma_d^2 x_d x_d' + \left( 1 + \frac{(x-x')^2}{2\alpha \sigma_l^2}\right)^{-\alpha}~.
    \end{split}
\end{equation*}

Here, \(x\) are our input parameters, while \(\sigma_d\), \(\sigma_L\) and \(\alpha\) are hyperparameters of the covariance function that can be estimated by maximizing the marginal log-likelihood.
To validate the approach, we perform a 10-fold cross-validation, also comparing alternative covariance functions (a radial basis function covariance function and a Matérn class covariance function; in sum and product combination with a linear covariance function), each time estimating the covariance function's hyperparameters using maximum marginal log-likelihood. For testing purposes, we condition the model with the best performance (selecting the covariance function and its hyperparameters from the best fold) on all train data and find a model performance of \(R^2 = 0.91\), which is comparable to the state-of-the-art heuristic model \cite{agrawal2014}. For use in the Bayesian Inference module, we condition the model on all available data afterward keeping the hyperparameters fixed.   

\subsection{Bayesian Inference Module}
In the (second) Bayesian Inference module, the GP's prediction is used as a prior for estimating a posterior distribution over the fatigue strength parameters (\(\mu_L, \sigma_L\)) using experimental data. 

The likelihood in the Bayesian Inference module is defined based on the experimental setup. We know that the failure probability of an experiment at load \(l\) follows the unknown cumulative log-normal distribution, leading to:
\begin{equation*}
    p(\text{outcome}(l) = \text{failure} \vert l) = \Phi_{\mu_L, \sigma_L}(l)~.
\end{equation*}
As the experiment is either a failure or a runout, we find:
\begin{equation*}
    \begin{split}
        p(\text{outcome}(l)& = \text{runout} \vert l) \\ 
        &= 1 -  p(\text{outcome}(l) = \text{failure} \vert l) = \\
        &= 1 - \Phi_{\mu_L, \sigma_L}(l)~.
    \end{split}
\end{equation*}
To gain the likelihood \(e\) of the outcome of an experimental series, we take the product of the individual probabilities, so
\begin{equation*}
    \begin{split}
        e(\text{outcome} &\vert \mu_L, \sigma_L)  \\
        &= \prod_i \Phi_{\mu_L, \sigma_L}(l_i) \cdot \prod_j \left(1- \Phi_{\mu_L, \sigma_L}(l_j) \right) ~,
    \end{split}
\end{equation*}
where \(i\) indexes the failures and \(j\) the runouts.
Given priors \(p(\cdot)\)  for the fatigue strength parameters (\(\mu_L, \sigma_L\)), we are able to define a joint (unnormalized) posterior distribution:
\begin{equation}
    \begin{split}
        &g(\mu_L, \sigma_L | \text{outcome}) = \\
        p(\mu_L&) \cdot p(\sigma_L) \cdot \left(
        \prod_i \Phi_{\mu_L, \sigma_L}(l_i) \cdot \prod_j \left(1- \Phi_{\mu_L, \sigma_L}(l_j) \right)
        \right)
    \end{split}
    \label{eq:posterior}
\end{equation}
By \(p(\cdot)\), we indicate a prior: for the mean fatigue strength \(\mu_L\), this is the GP's prediction; for the standard deviation, we use a fixed value, defined by the process engineers, as there is no practical interest in its estimation. Using a uniform positive or a gamma distribution is also possible.

This posterior joins the knowledge about the experimental setup with the knowledge already incorporated in the GP: the correlation of the fatigue strength for different materials.
It offers three options: the MAP estimation of the fatigue strength parameters (\(\mu_L, \sigma_L\)), the definition of Active Learning-style acquisition functions, and the calculation of the standard deviation of the posterior measuring the confidence of the MAP estimates.

\subsubsection{Maximum a Posteriori Estimation of the Fatigue Strength Parameters}
Given the expression \(g(\mu_L, \sigma_L)\) for the posterior distribution over the fatigue strength parameters, a straightforward approach is to estimate the most probable parameters by maximization. Thus, we find:
\begin{equation*}
    \hat{\mu}_L, \hat{\sigma}_L =\mathop{\arg\max}\limits_{\mu_L, \sigma_L} g(\mu_L, \sigma_L)~.
\end{equation*}

\subsubsection{Confidence of the Maximum a Posteriori Estimates}
Assuming that the prior over the distribution parameters is valid, we define a measure of the confidence of the found estimates using the standard deviation of the posterior: \(\text{Std}(g(\mu_L, \sigma_L))~.\)

We implement its approximation via numerical integration. Therefore, we sample a grid of 100,000 equidistant points at a support of plus/minus two predictive standard deviations from the GP's predictive mean. Given these evaluations of the posterior, we approximate the standard deviation numerically.

This rather small support has shown to work well in practice, as usually the GP's prediction is a lot broader than the posterior \(g(\mu_L, \sigma_L)\).

\subsubsection{Acquisition Functions}
In this work, we examine two potentially useful acquisition functions, identifying their robustness. In practice, their recommendations can be directly used in the lab to guide the process engineers when composing an experimental series.
\paragraph{Probability-weighted Entropy}
The probability-weighted entropy acquisition function joins two properties of an experimental outcome at a new load \(l\). On the one hand, there is the probability of the expected outcome; on the other hand, there is the experiment's impact on the posterior distribution.
Intuitively, we want to find the experiment that most probably carries the most information on the distribution parameters.
To express this mathematically, we make use of the current MAP estimate of the distribution parameters (\(\hat{\mu}_L, \hat{\sigma}_L\)). With it, we calculate the failure probability \(\Phi_{\hat{\mu}_L, \hat{\sigma}_L}(l)\) of an experiment at load \(l\) arising from the current state of data. 
Furthermore, we calculate the impact of the experimental outcome on the current posterior by its entropy \(H \left(g(\mu_L, \sigma_L \vert \text{outcome}(l), l)\right)\).
As a main goal, we want to minimize the entropy of the expected posterior distribution, when adding an experiment at load \(l\).

Combining both metrics, we find the next load \(l^\star\) by maximizing the following acquisition function:
\begin{equation}
    \begin{split}
        l^\star &= \mathop{\arg\max}\limits_l \\
        &\left(- H(g(\mu_L, \sigma_L \vert  {\text{outcome}(l) = \text{failure}}, l))\right) \cdot \Phi_{\hat{\mu}_L, \hat{\sigma}_L}(l)  \\
        + &\left(- H(g(\mu_L, \sigma_L \vert {\text{outcome}(l) = \text{runout}}, l)) \right) \\ 
        &~~~~~~~~~~~~~~~~~~~~~~~~~~~~~~~~~~~~~~~~~~~~~~~~~~~~~~~~~~~~~\cdot (1- \Phi_{\hat{\mu}_L, \hat{\sigma}_L}(l))~.
    \end{split}
\end{equation}

The calculation of information-based acquisitions often leads to a fast convergence but, as in standard Bayesian optimization, is numerically complex \cite{Hennig2012EntropySF,hernandez2014predictive,Wang2017MaxvalueES}. In our case, we approximate the entropy via a discrete set of data points, as the posterior follows a non-standard distribution. Therefore, we randomly sample 10,000 points from the points used for estimating the standard deviation of the posterior, using the current posterior distribution as a weight function.

\paragraph{Current MAP Estimate}
To minimize the computing time, we define a simpler approximation of the loads found by the probability-weighted entropy acquisition function: the current MAP estimate of the mean fatigue strength \(\hat{\mu}_L\). We expect this approximation to gain correctness in the course of experimentation, as the most informative load will become the one with the least unclear outcome, which is the mean fatigue strength. 

Therefore, we create the following acquisition function:
\begin{equation}
    l^\star = \mathop{\arg\max}\limits_{\mu_L} g(\mu_L, \sigma_L)~.
\end{equation}

\section{Robustness in Fatigue Strength Estimation}
In preparation for a feasibility study, we found several robustness challenges that our approach has to face. This paper describes two of them: prior misspecification and discretization of the specified loads.
\subsection{Prior Misspecification}
The prediction of the GP, used as a prior in the Bayesian Inference-module, is questionable for out-of-distribution samples, e.g., using cast iron instead of stainless steel. The model extrapolates by the constant mean and variance of the GP prior, but a detailed study of the correctness of this extrapolation behavior is speculative.

Instead, we study the convergence behavior of the different acquisition functions for a misspecified prior. For applying the Bayesian Inference approach, knowing how fast the experimental data can correct a wrong prior is imperative.

\subsection{Discretization of the Specified Loads}
Both the staircase method and our approach work on continuous load space and thus recommend experiments from a continuous load space. The real-life application requires at least the discretization of the loads to integer values or even broader fixed load levels. Therefore, we study the convergence behavior for differently discretized load values.

\section{Experiments}
\begin{figure*}[t]
    \begin{minipage}{0.29\textwidth}
        \raggedleft
        prior width \(10^{0.1}\)~N
    \end{minipage}
    \begin{minipage}{0.31\textwidth}
        \raggedleft
        prior width \(10^{1}\)~N
    \end{minipage}
    \begin{minipage}{0.319\textwidth}
        \raggedleft
        prior width \(10^{10}\)~N
    \end{minipage}
    
    \begin{minipage}{0.02\textwidth}
        \begin{turn}{90} 
            mean misspec. -75~\%
        \end{turn}
    \end{minipage}
    \hfill
    \begin{minipage}{0.319\textwidth}
        \includegraphics{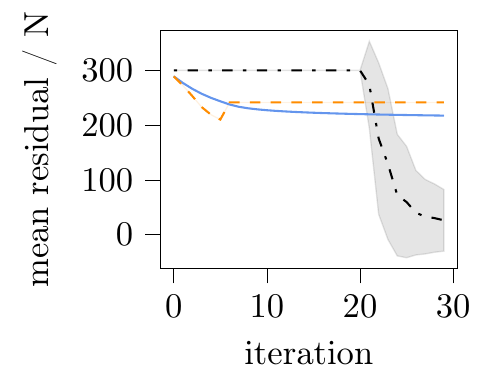}
    \end{minipage}
    \begin{minipage}{0.319\textwidth}
        \includegraphics{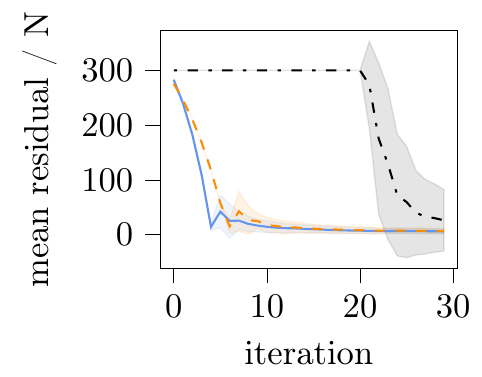}
    \end{minipage}
    \begin{minipage}{0.319\textwidth}
        \includegraphics{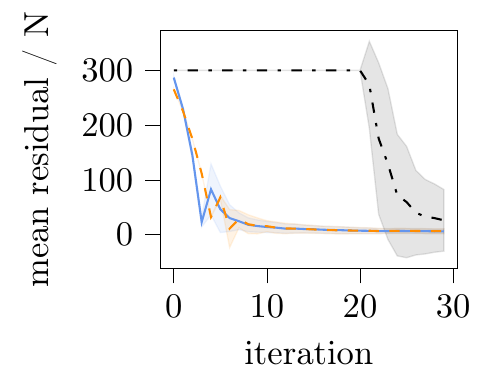}
    \end{minipage}
    
    \begin{minipage}{0.02\textwidth}
        \begin{turn}{90} 
            mean misspec. 0~\%
        \end{turn}
    \end{minipage}
    \hfill
    \begin{minipage}{0.319\textwidth}
        \includegraphics{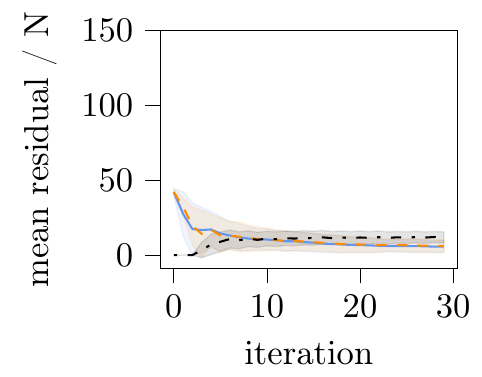}
    \end{minipage}
    \begin{minipage}{0.319\textwidth}
        \includegraphics{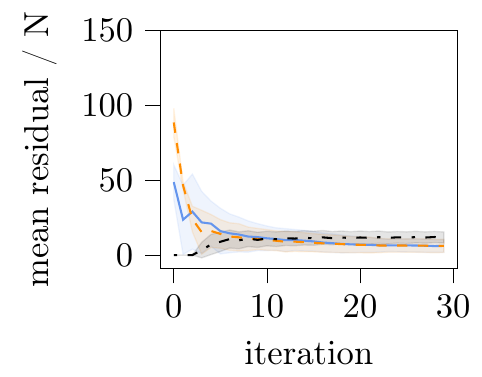}
    \end{minipage}
    \begin{minipage}{0.319\textwidth}
        \includegraphics{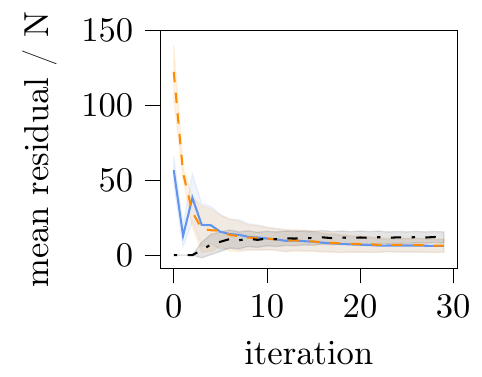}
    \end{minipage}
    
    \begin{minipage}{0.02\textwidth}
        \begin{turn}{90} 
            mean misspec. 75~\%
        \end{turn}
    \end{minipage}
    \hfill
    \begin{minipage}{0.319\textwidth}
        \includegraphics{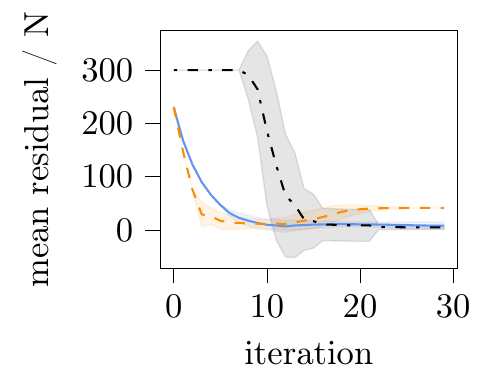}
    \end{minipage}
    \begin{minipage}{0.319\textwidth}
        \includegraphics{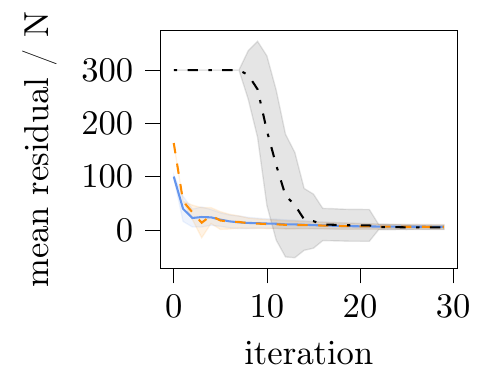}
    \end{minipage}
    \begin{minipage}{0.319\textwidth}
        \includegraphics{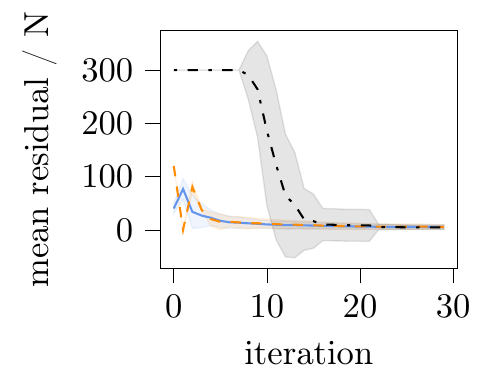}
    \end{minipage}
    \caption{Residuals \(\vert \mu_L - \hat{\mu}_L \vert\) of the estimated mean fatigue strength for different values of prior misspecification. \textcolor{blue}{-}: entropy acquisition; \textcolor{orange}{-~-}: MAP acquisition; \textcolor{black}{-\(\cdot\)-}: staircase method. The shaded areas indicate one standard deviation of the residual.}
    \label{fig:results_misspecification}
\end{figure*}
\begin{figure*}
    \begin{minipage}[c]{0.31\textwidth}
        \centering
            \subcaptionbox{Entropy acquisition.}{\includegraphics{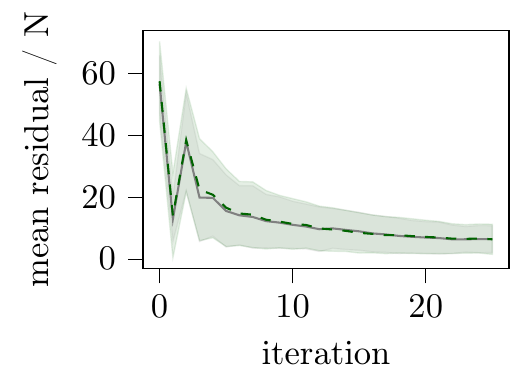}}
    \end{minipage}
    \hfill
    \begin{minipage}[c]{0.31\textwidth}
        \centering
            \subcaptionbox{MAP acquisition.}{\includegraphics{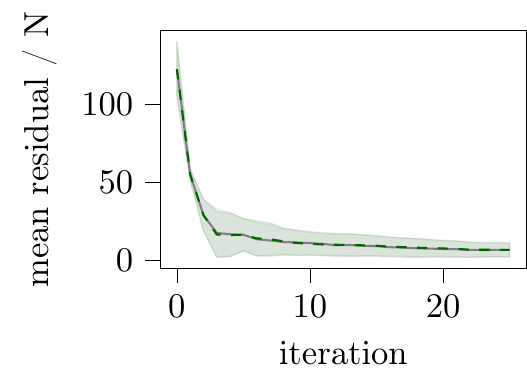}}
    \end{minipage}
    \hfill
    \begin{minipage}[c]{0.31\textwidth}
        \centering
            \subcaptionbox{Staircase method.}{\includegraphics{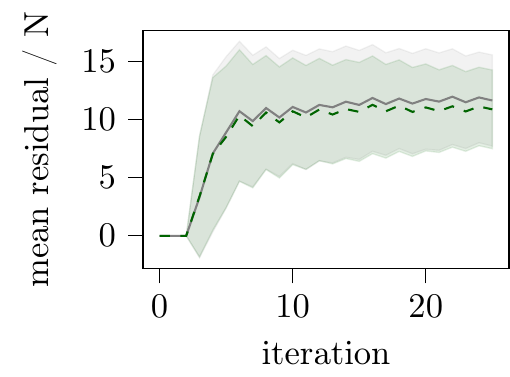}}
    \end{minipage}
    \caption{Convergence behavior for different discretizations, measured by residuals of the mean fatigue strength \(\vert \mu_L - \hat{\mu}_L \vert\). \\ \textcolor{black}{-}: No discretization. \textcolor{darkgreen}{-~-}: Discretization to multiples of ten.}
    \label{fig:rounding_convergence}
\end{figure*}
The basis for our experimental setup is a simulator for fatigue experiments that returns the outcome (failure or runout) for a specified load and given ground truth parameters (\(\mu_L, \sigma_L\)). For \(\mu_L\) we use a value of 400~N, \(\sigma_L\) is fixed for all experiments to a value of \(10^{0.03}\)~N. With these parameters given, simulating an experiment at load \(l\) means sampling its outcome according to the failure probability of an experiment \(\Phi_{\mu_L, \sigma_L}(l)\).

For each acquisition function and the staircase benchmark, we perform 100 runs to observe the convergence behavior at the different robustness requirements. For studying prior misspecification, we use 30 iterations; for the effect of discretization, we use 25. 
In the case of our acquisition functions and for MAP estimation, we use a multi-start approach of scipy's implementation of the Nelder-Mead simplex algorithms for optimization \cite{NelderMead}\footnote{\url{https://github.com/scipy/scipy/blob/main/scipy/optimize/_optimize.py}}.

For the prior misspecification, we distinguish between misspecification of the mean of the prior by \(\lbrace-75, 0, 75\rbrace\)~\% of the ground truth value and the width of the prior that varies standard deviations of \(\lbrace10^{0.1}, 10^1, 10^{10} \rbrace\)~N. For the staircase method, it is only possible to study a misspecified prior mean, as it corresponds to this experimental procedure's initial load \(L_{ini}\). There is no directly comparable hyperparameter to the prior width, but we choose the step size parameter \(d = {\sigma_L}\), according to the standard \cite{DIN}. This results in load steps from the set \(\lbrace 325, 348, 373, 400, 429, 460, 493\rbrace\)~N.  

Regarding the discretization requirement, we consider discretization factors from the set \(\lbrace \mbox{None}, -1\rbrace\), where \(\mbox{None}\) indicates no discretization, and -1 a discretization to multiples of ten (this one is common in the lab). 

\subsection{Convergence Behavior at Prior Misspecification}
Figure~\ref{fig:results_misspecification} shows the convergence behavior of the different approaches for a misspecified prior. 

It is evident that the staircase method is susceptible to a misspecified prior mean of \(\mu_L\). At the same time, it is insensitive to the width of the prior (as it is not a parameter of the method). Additionally, the sensitivity is not symmetric: in the case of underestimation, the staircase method converges slower than if the mean fatigue strength is overestimated - this is due to the load steps in the staircase method depending on the initial estimate of the mean fatigue strength \(\hat{\mu}_L \approx L_{ini}\).
The small residuals in the first iterations are due to the fact that in these cases, the experimental series does not fulfill the analysis requirements. Therefore, we estimate \(\hat{\mu}_L = L_{\text{ini}}\). Interesting is the fact that the staircase method does not converge to a residual of zero, but a slightly higher value around 10~N. A possible reason for this behavior is the applied stepsize: the width of the steps does not allow for a more precise estimation of the mean fatigue strength \(\hat{\mu}_L\). This convergence issue also applies to the experiments examining the discretization of loads.

In the case of no mean misspecification, the Bayesian Inference-based acquisition functions show their superiority over the staircase method: they fastly converge to a small residual, while the residual of the staircase method grows to a fixed offset because some of its runs did not converge at all. Additionally, we observe that the effect of the prior width is small: after ten iterations, the results become nearly identical.

Also, for a wide prior (i.e., prior width = \(10^{10}\)~N), the staircase method converges later than the other acquisition functions, especially if the initial estimate of the mean fatigue strength is misspecified.

In case of a very tight prior (i.e., prior width = \(10^{0.1}\)~N), the Bayesian Inference-based acquisition functions are sensitive to misspecification of the mean: in the case of an overestimated mean, both methods fail, while in the case of underestimation, only the MAP acquisition diverges. The divergence is because the prior has no probability mass in the area of the ground truth value, so the algorithms estimate some wrong posterior due to numerical reasons and diverge.

Apart from this case of a very tight misspecified prior, the entropy search and the MAP acquisition function behave similarly, thus confirming our assumptions on the approximation quality when using the MAP acquisition function.

\subsection{Convergence Behavior when Discretizing Loads}
\begin{figure}
    \centering    \includegraphics{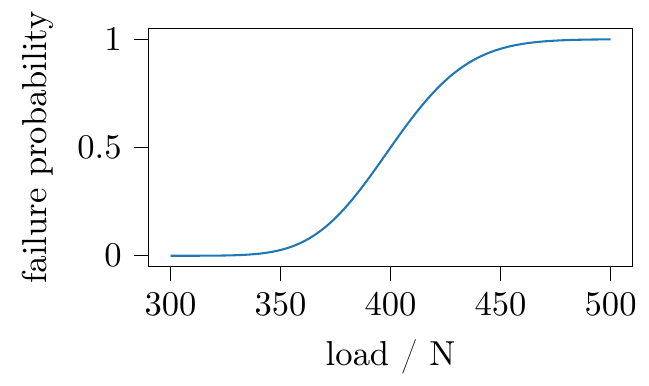}
    \caption{Applied failure probability \(\Phi_{\mu_L, \sigma_L(l)}\) in the experiments.}
    \label{fig:failure_probability}
\end{figure}

In figure~\ref{fig:rounding_convergence} we show the results for different load discretizations in case of a wide (i.e., standard deviation of \(10^{10}\)~N), non-misspecified prior. 
For all acquisitions, we do not observe any significant differences, indicating that all methods are robust against a discretization of the recommended loads. 
Mainly this is due to the steepness of the (realistic) failure probability curve (see figure~\ref{fig:failure_probability}) being too flat to impact the convergence.
As this discretization is the worst case at deployment time, we conclude that all methods are robust enough to face errors due to discretization. 

\subsection{Consequences for Deployment}
For deployment, this study leads to a simple conclusion: as long as the prior width is not extremely underestimated, the Bayesian Inference-based methods converge faster and are more robust against misspecifications than the staircase method.

\section{Limitations}
This approach was designed for the fatigue strength estimation of stainless steels. Two factors limit its application:
The quality of the GP's prediction and the quality of the posterior.

The robustness study shows that the Bayesian Inference module can correct a wrong prior as long as it is wide enough. The quality of the prior impacts the convergence behavior - the better the prior, the fewer experiments must be taken out.
A first step to improve the GP model (and thus the quality of the prior) is to expand the training data set, e.g., by using the open-source NIMS dataset \cite{NIMS_fatigue_data}.
If the GP's prediction is questionable, e.g., when inferring the mean fatigue strength for a stainless steel with heat treatment (a currently constant factor), it is advisable to replace the prior in the Bayesian Inference module with a broad one. Consequently, more experiments have to be taken out, but a good convergence to the ground truth value is still given.

The definition of the likelihood mainly influences the quality of the posterior in the Bayesian Inference module. Here, we follow the current standard and assume that the failure probability follows a log-normal distribution \cite{DIN}. To the authors' knowledge, this assumption is correct for all materials that undergo fatigue testing. Therefore, we expect that the approach generalizes well in its application area.

\section{Conclusion}
This paper describes a two-stage material characterization approach consisting of a GP and a Bayesian Inference module. This approach has several advantages over the state-of-the-art standardized method, especially being free of hyperparameters at application time. As the practical use of the approach demands its robustness against real-life requirements, we examined the following two issues: a misspecified prior and a discretization of the specified loads during the experimental procedure. Our results prove the robustness of our approach and its applicability in the lab, potentially reducing the number of costly experiments.

\section{Acknowledgements}
This work was partly developed by the Fraunhofer Center for Machine Learning within
the Fraunhofer Cluster for Cognitive Internet Technologies and has been partly funded by the Federal Ministry of 
Education and Research of Germany as part of the competence center for 
machine learning ML2R (01IS18038B).

\bibliography{literature.bib}

\appendix

\end{document}